\documentclass[10pt,a4paper]{article}

\usepackage[T1]{fontenc}
\usepackage[utf8]{inputenc}
\usepackage{newtxtext,newtxmath}
\usepackage[a4paper,margin=27mm]{geometry}
\usepackage{microtype}
\usepackage{graphicx}
\usepackage{booktabs}
\usepackage{array}
\usepackage{amsmath}
\usepackage{siunitx}
\usepackage[numbers,sort&compress]{natbib}
\usepackage{url}
\usepackage[hidelinks]{hyperref}
\newcommand{\doi}[1]{\href{https://doi.org/#1}{\nolinkurl{https://doi.org/#1}}}
\usepackage{authblk}
\usepackage{caption}
\usepackage{float}
\usepackage{placeins}

\title{Markovian Circuit Tracing: Ground-Truth Tests of Transformer State Abstractions}
\author[1,2]{Abdullah X\thanks{Corresponding author: \href{mailto:abdullah@projectawareai.org}{abdullah@projectawareai.org}}}
\affil[1]{Project AWARE, New Delhi, India}
\affil[2]{Zephara AI, New Delhi, India}
\date{}

\begin{document}
\maketitle

\begin{abstract}
Mechanistic interpretability needs tests that separate accessible information from a causally useful internal abstraction. We introduce Markovian Circuit Tracing (MCT), a controlled benchmark for testing whether transformer activations support a compact state abstraction with recoverable transition structure. We train small causal transformers on sequences from four-state hidden Markov models. The experiment fixes the latent transition law and varies emission observability. The canonical artifact contains 15 runs across five fixed seeds. Every model reaches validation loss within 0.02 nats per token of the exact Bayes predictor. Under easy observability, held-out mean rowwise Kullback-Leibler (KL) divergence from the true transition matrix is 0.114. The easy shuffled control gives 0.379. Under medium observability, recovered transition KL is 0.221. The medium shuffled control gives 0.355. Recovery collapses to control performance under hard observability. Natural activation swaps and state forcing show the same decline in causal specificity. New transition-specific controls refine this result. Untrained activations yield similar transition recovery in the easy regime. In medium, trained activations improve over the untrained control by 0.049 KL on average. Probabilistic Markov-order tests show a clear first-order gain in easy. Medium shows a smaller gain. Hard shows no meaningful gain. A second latent transition law reproduces the same observability-dependent pattern. The belief-probe control fails to support learned specificity. The fixed sparse autoencoder fails to improve transition recovery. MCT therefore tests when a state-machine interpretation earns support and when the evidence rejects it.
\end{abstract}


\section{Introduction}

Mechanistic interpretability asks how a trained neural network implements its behavior. Much of the modern literature isolates components and activation pathways that mediate specific model behaviors \citep{olah2020zoom,elhage2021framework,meng2022locating,conmy2023automated}. Sequential prediction raises a complementary problem because a model must update information across time. A useful explanation describes that evolving information and tests whether the proposed description predicts model behavior.

A state abstraction offers one route. If a model repeatedly enters internal regions with stable predictive meaning, we can test whether the trajectory through those regions admits an approximate transition law. This description can compress a long activation sequence into a smaller stochastic process. Natural-language settings provide no exact internal-state ground truth, which makes evaluation difficult. Probe accuracy creates a second difficulty because a probe can reconstruct information that the model never learned as a distinct internal variable \citep{hewitt2019designing,belinkov2022probing}.

Hidden Markov Models (HMMs) provide exact ground truth for this question \citep{baum1966statistical,rabiner1989tutorial}. The experimenter knows the latent-state sequence exactly. The experimenter also knows the transition law. The HMM supplies exact Bayesian predictive beliefs. It also supplies exact state-conditioned token distributions. We can therefore test an activation-derived abstraction against the generating process and against interventions with exact targets.

Recent work makes this benchmark especially relevant. Shai et al. show that transformer residual streams can represent predictive belief-state geometry for hidden generative processes \citep{shai2024belief}. Tamkin et al. impose sparse discrete codebook features and demonstrate state-like control on a finite-state-machine task \citep{tamkin2024codebook}. Torre et al. reverse-engineer recurrent networks that emulate HMM dynamics and identify continuous mechanisms that support stochastic state transitions \citep{torre2025hmm}. MCT asks a narrower empirical question. It begins with ordinary trained transformer activations. Calibration data determine a post-hoc discrete abstraction. Untouched trajectories then test that abstraction. MCT also measures whether observability controls the validity of the resulting state-machine description.

\begin{figure}[t]
\centering
\includegraphics[width=\textwidth]{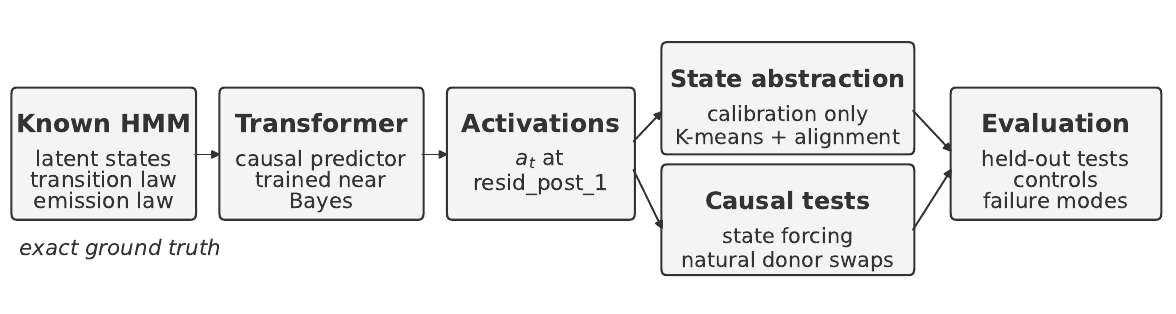}
\caption{Markovian Circuit Tracing benchmark. Calibration data determine the state abstraction and intervention centroids. Untouched evaluation sequences determine transition and causal results}
\label{fig:pipeline}
\end{figure}

We evaluate MCT through held-out transition recovery and causal interventions. State forcing replaces a residual activation with a calibration-derived state centroid and compares the output with the exact HMM target. A natural-donor swap test compares donors from the same recovered state with donors from a different recovered state. Figure~\ref{fig:pipeline} summarizes the benchmark.

The final design also separates transition structure that training creates from structure already available through the architecture or visible history. An identically configured untrained transformer supplies a transition-specific control. Explicit four-token history features provide a second control. We supplement the original Markov-order accuracy diagnostic with held-out probabilistic negative log likelihood (NLL). We also repeat the transition analysis under a second latent transition matrix as a post-hoc robustness check.

Our evidence supports a conditional claim. Easy observability yields strong transition recovery relative to shuffled and random controls. The untrained transformer already exposes nearly the same easy-regime transition structure. Medium observability produces a clear trained-over-untrained transition advantage. Hard observability removes the transition advantage and causal state selectivity. MCT therefore functions as a diagnostic for the evidential strength of a state abstraction. A first-order Markov description of the full transformer computation remains a stronger, unsupported interpretation.

\section{Background and related work}

\subsection{Mechanistic interpretability and causal intervention}

Circuit-based interpretability studies internal structures that mediate model behavior \citep{olah2020zoom,elhage2021framework}. Activation patching and causal tracing test whether changing a candidate representation changes the model output in the predicted direction \citep{meng2022locating}. Causal scrubbing raises the standard further by asking whether interventions preserve behavior when the hypothesized high-level variable remains fixed \citep{chan2022causal}. Formal work on causal abstraction asks whether neural variables realize the intervention structure of a higher-level model \citep{geiger2021causal}.

MCT applies this causal principle in a setting where the experiment specifies the higher-level stochastic process exactly. The benchmark evaluates transition structure first. It then checks whether state-conditioned interventions produce the expected predictive effect.

\subsection{Belief-state geometry and predictive states}

Linear probes test whether a representation exposes information to a simple decoder \citep{alain2016understanding,hewitt2019designing,belinkov2022probing}. Shai et al. provide direct evidence that transformer residual streams can encode the geometry of Bayesian belief states under controlled hidden processes \citep{shai2024belief}. Their result motivates the predictive-belief diagnostic in MCT. MCT adds a distinct test: a proposed discrete abstraction must yield held-out transition structure and causal state specificity. The benchmark also compares trained activations against an untrained transformer. Accessibility alone provides insufficient evidence of learning.

Predictive-state representations define state through information about future observations \citep{jaeger2000observable,littman2001predictive,boots2011closing}. This view matters under partial observability because a transformer cannot generally infer the sampled HMM state exactly from the visible prefix. The Bayesian predictive belief provides the sufficient statistic available to an optimal causal predictor.

\subsection{Discrete internal states}

Earlier work extracted finite-state structure from recurrent networks and related sequence models \citep{cleeremans1989finite,omlin1996extraction,weiss2018practical}. Tamkin et al. train neural networks with discrete codebook features and show causal control on a finite-state-machine dataset \citep{tamkin2024codebook}. Their method changes the representation during training through quantization. MCT leaves the trained transformer unchanged and evaluates a post-hoc abstraction against known stochastic dynamics.

Torre et al. study recurrent networks that emulate HMMs and uncover continuous dynamical mechanisms that reproduce stochastic state transitions \citep{torre2025hmm}. Their analysis shows that a discrete generating process can still produce continuous neural states. MCT addresses the complementary question of when a coarse discrete abstraction remains empirically useful for a transformer and where that abstraction fails.

\subsection{Sparse representations}

Sparse coding and sparse autoencoders (SAEs) seek feature bases intended to simplify neural representations \citep{olshausen1997sparse,elhage2022superposition,huben2024sparse}. MCT includes a fixed sparse-autoencoder comparison to test whether sparsification improves state separation. The confirmatory experiment treats the fixed SAE as an ablation and reports its failure to improve transition recovery.

\section{Problem setting and Markovian Circuit Tracing}

\subsection{Temporal semantics}

Each sequence comes from a finite HMM with hidden state $s_t\in\{1,\ldots,K\}$ and visible token $x_t\in\{1,\ldots,V\}$. At time $t$, state $s_t$ emits $x_t$ according to emission matrix $E$. The process then transitions to $s_{t+1}$ according to transition matrix $T$:
\begin{equation}
P(x_t=v\mid s_t=i)=E_{iv}, \qquad
P(s_{t+1}=j\mid s_t=i)=T_{ij}.
\end{equation}

The causal language model uses a beginning-of-sequence token at position zero. At model position $t$, its input contains exactly $x_{<t}$ and its target is $x_t$. The predictive belief therefore matches the activation at that position:
\begin{equation}
 b^-_t(i)=P(s_t=i\mid x_{<t}).
\end{equation}
The Bayes-optimal token distribution is
\begin{equation}
 P(x_t\mid x_{<t})=b^-_t E.
\end{equation}
After the process emits $x_t$, the filtering posterior becomes
\begin{equation}
 b^+_t(i)=P(s_t=i\mid x_{\le t}).
\end{equation}
MCT uses $b^-_t$ as the primary belief target because the model has access to the same information when it predicts $x_t$.

This alignment also determines the intervention target. If an intervention represents $s_t=k$ at the same model position, the exact target distribution is
\begin{equation}
 P(x_t\mid s_t=k)=E_{k,:}.
 \label{eq:forcingtarget}
\end{equation}
The quantity $T_{k,:}E$ describes the next position, $P(x_{t+1}\mid s_t=k)$, so MCT excludes it from the same-position forcing target.

\subsection{Activation-derived state abstraction}

Let $a_t\in\mathbb{R}^d$ denote the chosen transformer activation. The canonical experiment uses \texttt{resid\_post\_1}. MCT fits K-means with $K=4$ on calibration activations. Hungarian matching then assigns each cluster to an HMM state using calibration labels. MCT freezes the geometry and mapping before evaluating any final sequence.

The alignment uses oracle calibration labels for benchmark evaluation. K-means learns cluster geometry without evaluation labels. Hungarian matching establishes correspondence with named HMM states. The procedure therefore tests the quality of a state abstraction under controlled ground truth. Automatic recovery of named states in an unlabeled deployment setting lies outside this benchmark.

Applying the frozen abstraction to an evaluation trajectory produces recovered states $z_t$. MCT estimates
\begin{equation}
 \widehat T_{ij}
 =\frac{\sum_t \mathbf{1}\{z_t=i,z_{t+1}=j\}}
 {\sum_t \mathbf{1}\{z_t=i\}},
\end{equation}
with a small additive smoothing constant. The sums include transitions only within each sequence, so sequence boundaries never create transitions. The primary transition metric is $K^{-1}\sum_i D_{\mathrm{KL}}(T_{i,:}\Vert\widehat T_{i,:})$.

\subsection{Evaluation criteria}

MCT treats predictive competence as a prerequisite. Predictive-belief and sampled-state probes describe representational recoverability. A separate transition metric evaluates dynamical recoverability on untouched sequences. Intervention tests evaluate causal relevance. We report these measurements separately because success on one leaves the others unresolved.

\section{Experimental design}

\subsection{Canonical observability-controlled HMMs}

The canonical benchmark uses four hidden states and six visible symbols. Every observability regime uses the same latent transition matrix,
\begin{equation}
T=\begin{bmatrix}
0.70&0.20&0.10&0.00\\
0.10&0.60&0.20&0.10\\
0.00&0.20&0.70&0.10\\
0.25&0.25&0.25&0.25
\end{bmatrix}.
\end{equation}
The regimes differ only in emission overlap. Appendix~\ref{app:hmm} gives the emission matrices. This design isolates observability because the hidden dynamics remain fixed as visible evidence about the current state weakens.

\subsection{Model and split protocol}

Each canonical model is a two-layer causal transformer with hidden width 64 and four attention heads. The multilayer perceptron (MLP) width is 128. Sequences have length 40. Each run uses 1200 training sequences and 250 independent validation sequences. The batch size is 256, and the learning rate is $3\times10^{-4}$. Training runs for at least six epochs and stops once validation cross-entropy comes within 0.02 nats per token of the exact Bayes loss, subject to a 30-epoch cap.

A separate 800-sequence corpus supports interpretation. The sequence-level split assigns 60\% to calibration and 20\% to selection. The remaining 20\% forms untouched evaluation data. Calibration data determine probes and cluster geometry. They also determine cluster alignment and intervention centroids. The fixed SAE trains on calibration data. Final reported metrics use the evaluation split.

The canonical grid crosses each observability regime with five fixed seeds, producing 15 runs. Table~\ref{tab:protocol} lists the exact seeds and the remaining protocol settings. We froze the protocol after exploratory development and before executing the final 15-run evidence grid. The protocol had no external preregistration.

\begin{table}[t]
\centering
\caption{Canonical confirmatory protocol}
\label{tab:protocol}
\small
\begin{tabular}{p{0.32\textwidth}p{0.60\textwidth}}
\toprule
Component & Setting \\
\midrule
HMM & 4 states and 6 visible symbols; common transition matrix \\
Seeds & 7, 17, 29, 43, 71 \\
Transformer & 2 layers; $d_{\mathrm{model}}=64$; 4 heads; MLP width 128 \\
Training & 1200 train sequences; 250 validation sequences; sequence length 40 \\
Analysis split & 60\% calibration; 20\% selection; 20\% evaluation \\
Representation & \texttt{resid\_post\_1} \\
State abstraction & K-means with $K=4$ and calibration-only alignment \\
Forcing & Position 12; up to 64 evaluation sequences per seed \\
Activation swaps & Up to 48 recipient-donor pairs per seed \\
SAE & Width 128; 2 epochs; $\ell_1=0.003$; top-$k=16$ \\
\bottomrule
\end{tabular}
\end{table}

\subsection{Representation and transition controls}

A ridge regressor maps calibration activations to exact predictive beliefs. MCT reports held-out KL divergence and variance-weighted $R^2$. A deterministic ridge classifier supplies the sampled-state probe. The Bayes observable-state ceiling uses the most probable state under the exact predictive belief.

Belief-specificity controls use explicit token history and an identically configured untrained transformer. The token controls use history length $h\in\{1,2,4\}$. The untrained model passes through the same probe pipeline.

The original transition analysis uses shuffled recovered labels and independently random state labels as null processes. The revision adds an untrained-transition control. It applies the same K-means and calibration alignment procedure to \texttt{resid\_post\_1} activations from the untrained transformer. A four-token history control applies the same K-means procedure to causal one-hot history features. These additions are post-hoc robustness analyses and leave the frozen canonical claim ledger unchanged.

\subsection{Probabilistic Markov-order analysis}

The canonical artifact reports an accuracy-based Markov-order diagnostic. The revision adds held-out probabilistic NLL. Calibration recovered states estimate an order-0 distribution $P(z_t)$ and an order-1 transition law $P(z_t\mid z_{t-1})$. Calibration trajectories also estimate $P(z_t\mid z_{t-1},z_{t-2})$ for order 2. The evaluation trajectories provide the final NLL values. We report the order-0 to order-1 NLL reduction and the additional order-1 to order-2 reduction. Positive values indicate improved predictive fit.

\subsection{Causal interventions}

State forcing estimates recovered-state centroids from calibration activations. For target state $k$, MCT replaces the activation at position 12 with the aligned recovered centroid. It then compares the patched output with Eq.~\ref{eq:forcingtarget}. The primary controls use the unpatched model and a wrong recovered centroid. A mean-activation patch provides an additional control. A magnitude-matched random activation tests nonspecific perturbation. A deranged shuffled-label centroid tests label misalignment. A centroid grouped by sampled true states supplies an oracle comparison.

The natural-donor test uses evaluation activations directly. A donor either shares the recipient's recovered state or belongs to a different recovered state at the intervention position. MCT measures the KL change between the original output distribution and the patched output distribution. Under a useful state equivalence class, same-state donors produce smaller changes.

\subsection{Second transition-law robustness check}

The revision also evaluates a second latent transition matrix,
\begin{equation}
T_{\mathrm{cyclic}}=\begin{bmatrix}
0.58&0.32&0.07&0.03\\
0.05&0.58&0.32&0.05\\
0.03&0.07&0.58&0.32\\
0.32&0.05&0.05&0.58
\end{bmatrix}.
\label{eq:cyclicT}
\end{equation}
This post-hoc robustness grid keeps the canonical emission matrices and fixed seed set. It allows up to 60 training epochs so every model reaches the same 0.02-nat validation-gap target. We use this family only to test whether observability dependence survives a different latent transition law.

\subsection{Statistical analysis}

The canonical claim ledger uses paired comparisons across five fixed seeds. We report effect magnitudes first. Figure error bars show the mean plus or minus 1.96 standard errors across the five fixed seeds. We use two-sided paired $t$ tests as secondary summaries of seed-level differences. Five seeds provide a robustness check. They offer limited support for large-sample population inference. The post-hoc analyses use the same fixed seeds and preserve the original calibration and evaluation split logic.

\section{Results}

\subsection{Canonical models reach near-Bayes predictive performance}

All 15 canonical models reach the stopping condition. Mean held-out excess loss over the Bayes predictor is 0.0198 nats per token in easy and 0.0198 in medium. The hard-regime mean is 0.0190. The held-out mean excess losses remain close to the pre-specified validation tolerance.

\subsection{Belief decodability is high without learned specificity}

A linear probe reconstructs exact predictive beliefs from trained activations with held-out $R^2$ of 0.925 in easy and 0.849 in medium. Hard gives 0.777. The trained representation shows no reliable advantage over the untrained-transformer belief control. A four-token history baseline also matches or improves on trained-activation belief KL. The advantage of the history baseline is largest in medium and hard. Table~\ref{tab:belief} reports the values.

\begin{table}[t]
\centering
\caption{Predictive-belief recovery and specificity controls. Lower KL is better}
\label{tab:belief}
\small
\begin{tabular}{lrrrr}
\toprule
Regime & Trained KL & Four-token KL & Untrained KL & Trained $R^2$ \\
\midrule
Easy & 0.0291 & 0.0272 & 0.0319 & 0.925 \\
Medium & 0.0246 & 0.0114 & 0.0247 & 0.849 \\
Hard & 0.00294 & 0.000401 & 0.00293 & 0.777 \\
\bottomrule
\end{tabular}
\end{table}

The result supports belief accessibility. Learned-belief specificity requires additional evidence. This distinction agrees with the broader concern that probe success can reflect information already exposed by the input or architecture.

\subsection{Transition recovery depends on observability}

\begin{figure}[t]
\centering
\includegraphics[width=0.78\textwidth]{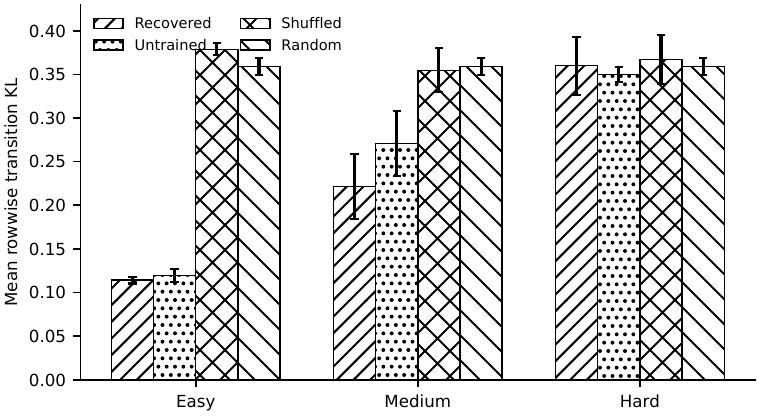}
\caption{Held-out transition recovery. Lower mean rowwise KL is better. Error bars show the mean plus or minus 1.96 standard errors across five fixed seeds. The untrained control uses the same clustering and calibration alignment pipeline}
\label{fig:transition}
\end{figure}

The frozen abstraction produces mean rowwise transition KL of 0.114 in easy and 0.221 in medium. Shuffled controls yield 0.379 and 0.355. Random states yield 0.359 in both regimes. Hard-observability recovery rises to 0.360, where shuffled and random controls give 0.367 and 0.359. Figure~\ref{fig:transition} shows the transition results together with the new untrained baseline.

The untrained control changes the interpretation of the easy regime. Its mean transition KL is 0.120. The untrained mean exceeds the trained value by 0.0055. The paired comparison gives $p=0.321$. The untrained architecture already exposes enough state-correlated structure for similar transition recovery under easy emissions. Transition recovery alone therefore provides no evidence of learning-specific structure in that regime.

Medium provides a different result. The untrained mean is 0.271. Training improves transition KL by 0.049 on average across paired seeds, with $p=0.0086$. The explicit four-token K-means control has mean transition KL of 0.964. Hard shows no trained advantage over the untrained representation. Table~\ref{tab:transitioncontrols} reports these controls.

\begin{table}[t]
\centering
\caption{Transition-specific controls added in this revision}
\label{tab:transitioncontrols}
\small
\begin{tabular}{lrrrrr}
\toprule
Regime & Trained & Untrained & Four-token history & Shuffled & Random \\
\midrule
Easy & 0.114 & 0.120 & 1.032 & 0.379 & 0.359 \\
Medium & 0.221 & 0.271 & 0.964 & 0.355 & 0.359 \\
Hard & 0.360 & 0.350 & 0.779 & 0.367 & 0.359 \\
\bottomrule
\end{tabular}
\end{table}

\subsection{Probabilistic Markov tests confirm the order effect}

\begin{figure}[t]
\centering
\includegraphics[width=0.70\textwidth]{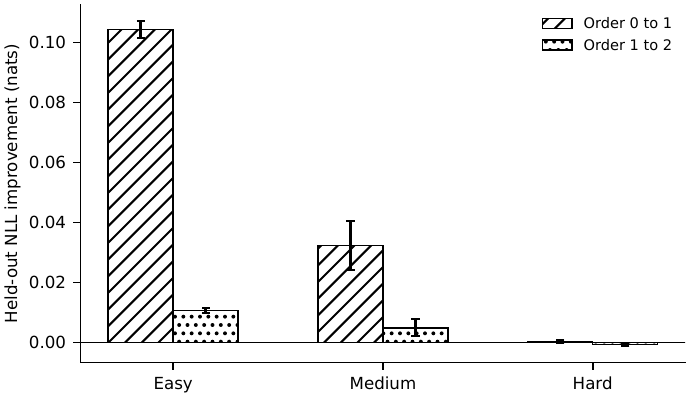}
\caption{Held-out probabilistic Markov-order gains. Positive values indicate lower NLL for the higher-order predictor. Error bars show the mean plus or minus 1.96 standard errors across five fixed seeds}
\label{fig:nll}
\end{figure}

The held-out NLL analysis gives a large order-0 to order-1 improvement in easy. Mean NLL falls from 1.371 to 1.266 nats, a gain of 0.104. Order 2 lowers NLL further to 1.256, which adds 0.0106. Medium shows a smaller first-order gain of 0.0323 and an additional order-2 gain of 0.0049. Hard shows essentially no first-order gain. Its mean improvement is 0.00035. Order 2 slightly worsens NLL on average. Figure~\ref{fig:nll} shows the seed-level uncertainty for both gains.

The probabilistic analysis therefore supports a useful first-order approximation under easy observability and a weaker approximation in medium. It provides no meaningful first-order evidence in hard. This conclusion matches the transition-recovery pattern without relying on argmax classification accuracy.

\subsection{Natural activation swaps show a shrinking causal effect}

\begin{figure}[t]
\centering
\includegraphics[width=0.58\textwidth]{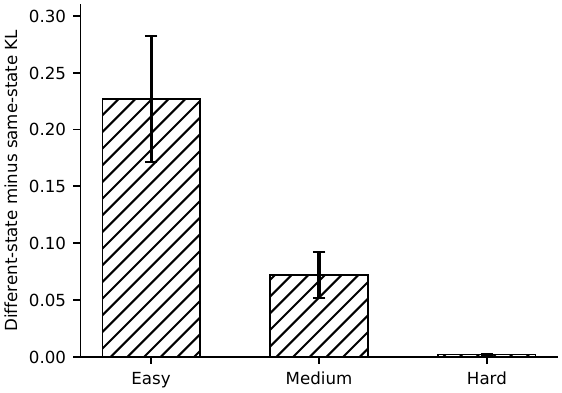}
\caption{Natural activation-swap effect. The metric subtracts same-state donor KL from different-state donor KL. Error bars show the mean plus or minus 1.96 standard errors across five fixed seeds}
\label{fig:swaps}
\end{figure}

Natural donor swaps supply a causal test that stays on the model's activation distribution. The different-state minus same-state output KL gap is 0.2269 in easy. Medium falls to 0.0722. Hard retains a positive mean gap of 0.00203, which is two orders of magnitude smaller than the easy effect. Figure~\ref{fig:swaps} summarizes these effects.

All five hard-regime seeds retain the positive sign. The paired test gives $p=0.00117$. The mean gap is only 0.00203, so MCT treats the hard result as a small effect.

\subsection{State forcing loses selectivity under hard observability}

The forcing experiment compares the model output with the exact emission distribution of the target state. In easy, unpatched KL is 0.785. The correct recovered centroid reduces it to 0.323. The wrong centroid gives 0.938. Medium shows the same qualitative selectivity. Correct forcing gives 0.188. The unpatched value is 0.385. The wrong-centroid value is 0.394.

\begin{figure}[t]
\centering
\includegraphics[width=0.70\textwidth]{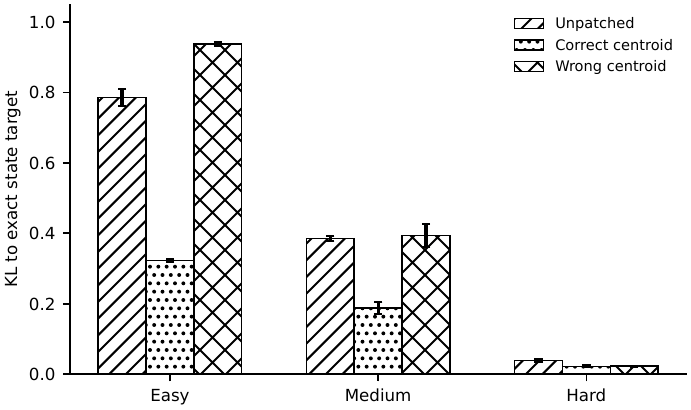}
\caption{State-forcing controls. Lower KL to the exact state-conditioned target is better. Error bars show the mean plus or minus 1.96 standard errors across five seed-level means}
\label{fig:forcing}
\end{figure}

Hard behaves differently. Correct forcing lowers KL from 0.0393 to 0.0223. The wrong centroid gives 0.0228. The state identity of the patch therefore provides no detectable selectivity in hard. Figure~\ref{fig:forcing} summarizes the seed-level means.

\subsection{The fixed sparse autoencoder fails to improve transition recovery}

The pre-specified SAE reaches transition KL of 0.117 in easy and 0.258 in medium. Hard rises to 0.416. The raw residual values are 0.114 in easy and 0.221 in medium. Hard gives 0.360. The SAE changes the easy-regime metric only slightly. It worsens the main transition metric in medium and hard. Table~\ref{tab:sae} reports the comparison. The fixed sparse representation therefore fails to improve MCT.

\begin{table}[t]
\centering
\caption{Raw residual and fixed-SAE transition recovery}
\label{tab:sae}
\small
\begin{tabular}{lrr}
\toprule
Regime & Raw residual KL & SAE KL \\
\midrule
Easy & 0.114 & 0.117 \\
Medium & 0.221 & 0.258 \\
Hard & 0.360 & 0.416 \\
\bottomrule
\end{tabular}
\end{table}

\subsection{A second latent transition law reproduces observability dependence}

The post-hoc cyclic family reaches the same validation-gap target in all 15 additional runs. Easy transition KL is 0.158. The shuffled value is 0.441, and the random value is 0.408. Medium transition KL is 0.324. Its shuffled value is 0.465, and its random value is 0.408. Hard rises to 0.466. The hard shuffled mean is 0.486. The random control is lower at 0.408. We therefore classify the hard cyclic result as a failure of robust transition recovery.

The probabilistic order test follows the same pattern. The order-0 to order-1 NLL gain is 0.0898 in easy and 0.0187 in medium. Hard falls to 0.00080. Table~\ref{tab:cyclic} summarizes the second-family results.

\begin{table}[t]
\centering
\caption{Post-hoc robustness under the second cyclic transition law}
\label{tab:cyclic}
\small
\begin{tabular}{lrrrr}
\toprule
Regime & Recovered KL & Shuffled KL & Random KL & NLL gain 0 to 1 \\
\midrule
Easy & 0.158 & 0.441 & 0.408 & 0.0898 \\
Medium & 0.324 & 0.465 & 0.408 & 0.0187 \\
Hard & 0.466 & 0.486 & 0.408 & 0.00080 \\
\bottomrule
\end{tabular}
\end{table}

The second transition family reproduces the main degradation with reduced observability. The repeated pattern under Eq.~\ref{eq:cyclicT} shows that the canonical transition matrix alone cannot explain the result.

\section{Discussion}

The final evidence supports MCT as a test of state abstractions. This benchmark provides no support for literal discrete neural states. The canonical trained models recover transition structure well relative to null sequences under easy observability. The new untrained control shows that easy-regime transition recovery lacks learning specificity. This result narrows the mechanistic claim and limits the evidential weight of transition recovery in the easy regime.

Medium observability provides the clearest training-specific dynamical result. Trained activations improve transition KL over the identically configured untrained transformer across every seed. They also retain causal swap separation and selective forcing. The combined evidence therefore goes beyond simple decodability in medium. It links learned activation geometry with held-out dynamics and state-sensitive behavior.

Hard observability marks the failure regime. Canonical transition KL matches the random-state control. The trained representation also matches the untrained transition control. Probabilistic first-order gain disappears. State forcing loses state identity. Natural donor swaps retain only a very small effect. These results remain coherent with a smoother predictive representation whose geometry lacks four reusable discrete regions.

The comparison with Shai et al. helps clarify the contribution \citep{shai2024belief}. Their work establishes rich belief-state geometry in transformer residual streams. Our belief probes reproduce the broader accessibility phenomenon in a simpler controlled HMM setting. MCT then tests whether a coarse discrete partition supports held-out dynamics and causal interventions. The untrained controls show why that additional standard matters.

The comparison with codebook features provides a second distinction \citep{tamkin2024codebook}. Codebook models learn a discretized representation during training. MCT leaves the model unchanged and subjects a post-hoc partition to ground-truth transition tests. The negative SAE result shows that added representation structure alone fails to improve the benchmark.

Torre et al. provide a complementary warning against literal state interpretations \citep{torre2025hmm}. Their recurrent networks emulate discrete HMM behavior through continuous dynamical mechanisms. Our hard-observability failure reaches a compatible lesson from transformer activations: useful coarse states can disappear even when the external generator still has discrete latent dynamics.

The second transition family strengthens the observability conclusion. We classify this analysis as post-hoc. It changes the latent transition structure. The first-order NLL gain remains large in easy. The gain becomes smaller in medium and approaches zero in hard. The next robustness step is to test more transition families and model architectures.

The benchmark still has important limitations. The experiments use small two-layer transformers on synthetic four-state HMMs, which limits the benchmark scale. Future work must test larger models. K-means also receives the true number of latent states. Calibration labels establish correspondence with HMM states. This oracle information limits deployment realism.

The causal evaluation has a narrower scope. It examines one residual site at one fixed intervention position. Centroid forcing can move an activation into a low-density region. The natural-donor test reduces this concern by using observed activations. Five seeds provide limited support for frequentist inference. We therefore emphasize effect magnitudes and per-seed consistency over isolated p-values.

The evidence presented here supports no state-machine claim for natural-language frontier models. The HMM setting provides exact ground truth so that the benchmark can distinguish strong evidence from attractive interpretation. A natural-language extension requires an independently specified candidate state variable and final examples that remain untouched during abstraction fitting. It also requires controls that test whether visible context or architecture already explains the apparent state structure.

\section{Conclusion}

Markovian Circuit Tracing evaluates whether a transformer state abstraction survives held-out dynamical and causal tests. The canonical experiment shows strong transition recovery relative to shuffled and random controls under easy observability. The result weakens in medium and fails under hard observability. The new untrained-transition control shows that easy-regime transition structure lacks learning specificity. Medium retains a clear trained-over-untrained advantage. Held-out probabilistic NLL shows a first-order predictive gain in easy and medium. Hard shows no meaningful gain. Causal donor swaps and state forcing show the same observability-dependent decline. A second latent transition matrix reproduces the pattern. These results make MCT useful as a test that can support a proposed state-machine interpretation or weaken it when the evidence fails.

\section*{Statements and Declarations}

\noindent\textbf{Funding.} The submitted work received no external funding or grant support.

\noindent\textbf{Competing interests.} Abdullah X is Founder and President of Project AWARE and Founder and CEO of Zephara AI. Both organizations appear as affiliations. The author declares no other relevant financial or non-financial interests related to this work.

\noindent\textbf{Ethics approval.} Not applicable. The study uses only synthetic data, with no human participants or animals.

\noindent\textbf{Consent to participate.} Not applicable.

\noindent\textbf{Consent for publication.} Not applicable.

\noindent\textbf{Data availability.} The public repository contains the frozen canonical evidence in \texttt{results/v1/}. The repository stores the additional transition controls in \texttt{results/npl\_robustness/}. The same directory contains the probabilistic Markov-order analysis and the second transition-family results. All data are synthetic and reproducible from the supplied code. The submission corresponds to repository commit \texttt{eb619d16f085}.

\noindent\textbf{Code availability.} Code and reproduction materials are available at \url{https://github.com/abdullah-x-bd/markovian-circuit-tracing}. The submission corresponds to repository commit \texttt{eb619d16f085}. The original canonical artifact is version 0.2.0. The repository stores the post-hoc robustness analyses separately, leaving the frozen canonical evidence unchanged.

\noindent\textbf{Author contributions.} Abdullah X conceived and designed the study. He implemented the software and conducted the experiments. He performed the formal analysis and prepared the figures. He wrote and revised the manuscript.

\noindent\textbf{Preprint.} The author posted an earlier exploratory version as arXiv:2605.20824. The present manuscript replaces its experimental design and headline results with the completed canonical artifact and clearly labeled post-hoc robustness analyses.

\bibliographystyle{unsrtnat}
\bibliography{references}

@article{baum1966statistical,
  author={Baum, Leonard E. and Petrie, Ted},
  title={Statistical inference for probabilistic functions of finite state Markov chains},
  journal={The Annals of Mathematical Statistics},
  volume={37}, number={6}, pages={1554--1563}, year={1966},
  doi={10.1214/aoms/1177699147}
}

@article{rabiner1989tutorial,
  author={Rabiner, Lawrence R.},
  title={A tutorial on hidden Markov models and selected applications in speech recognition},
  journal={Proceedings of the IEEE}, volume={77}, number={2}, pages={257--286}, year={1989},
  doi={10.1109/5.18626}
}

@article{olah2020zoom,
  author={Olah, Chris and Cammarata, Nick and Schubert, Ludwig and Goh, Gabriel and Petrov, Michael and Carter, Shan},
  title={Zoom In: An Introduction to Circuits},
  journal={Distill}, volume={5}, number={3}, year={2020}, doi={10.23915/distill.00024.001}
}

@misc{elhage2021framework,
  author={Elhage, Nelson and Nanda, Neel and Olsson, Catherine and Henighan, Tom and Joseph, Nicholas and Mann, Ben and Askell, Amanda and Bai, Yuntao and Chen, Anna and Conerly, Tom and others},
  title={A Mathematical Framework for Transformer Circuits}, year={2021},
  howpublished={Transformer Circuits Thread}, url={https://transformer-circuits.pub/2021/framework/index.html}, note={Accessed 6 October 2026}
}

@inproceedings{meng2022locating,
  author={Meng, Kevin and Bau, David and Andonian, Alex and Belinkov, Yonatan},
  title={Locating and Editing Factual Associations in {GPT}},
  booktitle={Advances in Neural Information Processing Systems}, volume={35}, pages={17359--17372}, year={2022},
  doi={10.52202/068431-1262}
}

@inproceedings{conmy2023automated,
  author={Conmy, Arthur and Mavor-Parker, Augustine and Lynch, Aengus and Heimersheim, Stefan and Garriga-Alonso, Adri\`a},
  title={Towards Automated Circuit Discovery for Mechanistic Interpretability},
  booktitle={Advances in Neural Information Processing Systems}, volume={36}, pages={16318--16352}, year={2023},
  doi={10.52202/075280-0719}
}

@article{chan2022causal,
  author={Chan, Lawrence and Garriga-Alonso, Adri\`a and Goldowsky-Dill, Nicholas and Greenblatt, Ryan and Nitishinskaya, Jenny and Radhakrishnan, Ansh and Shlegeris, Buck and Thomas, Nate},
  title={Causal Scrubbing: A Method for Rigorously Testing Interpretability Hypotheses},
  journal={AI Alignment Forum}, year={2022},
  url={https://www.alignmentforum.org/posts/JvZhhzycHu2Yd57RN/causal-scrubbing-a-method-for-rigorously-testing},
  note={Accessed 6 October 2026}
}

@inproceedings{geiger2021causal,
  author={Geiger, Atticus and Lu, Hanson and Icard, Thomas and Potts, Christopher},
  title={Causal Abstractions of Neural Networks},
  booktitle={Advances in Neural Information Processing Systems}, volume={34}, pages={9574--9586}, year={2021},
  url={https://proceedings.neurips.cc/paper/2021/hash/4f5c422f4d49a5a807eda27434231040-Abstract.html}
}

@inproceedings{alain2016understanding,
  author={Alain, Guillaume and Bengio, Yoshua},
  title={Understanding Intermediate Layers Using Linear Classifier Probes},
  booktitle={5th International Conference on Learning Representations, ICLR 2017, Workshop Track Proceedings},
  year={2017}, url={https://openreview.net/forum?id=HJ4-rAVtl}
}

@inproceedings{hewitt2019designing,
  author={Hewitt, John and Liang, Percy},
  title={Designing and Interpreting Probes with Control Tasks},
  booktitle={Proceedings of the 2019 Conference on Empirical Methods in Natural Language Processing and the 9th International Joint Conference on Natural Language Processing},
  pages={2733--2743}, publisher={Association for Computational Linguistics}, year={2019}, doi={10.18653/v1/D19-1275}
}

@article{belinkov2022probing,
  author={Belinkov, Yonatan},
  title={Probing classifiers: Promises, shortcomings, and advances},
  journal={Computational Linguistics}, volume={48}, number={1}, pages={207--219}, year={2022}, doi={10.1162/coli_a_00422}
}

@article{cleeremans1989finite,
  author={Cleeremans, Axel and Servan-Schreiber, David and McClelland, James L.},
  title={Finite state automata and simple recurrent networks},
  journal={Neural Computation}, volume={1}, number={3}, pages={372--381}, year={1989}, doi={10.1162/neco.1989.1.3.372}
}

@article{omlin1996extraction,
  author={Omlin, Christian W. and Giles, C. Lee},
  title={Extraction of rules from discrete-time recurrent neural networks},
  journal={Neural Networks}, volume={9}, number={1}, pages={41--52}, year={1996}, doi={10.1016/0893-6080(95)00086-0}
}

@inproceedings{weiss2018practical,
  author={Weiss, Gail and Goldberg, Yoav and Yahav, Eran},
  title={Extracting Automata from Recurrent Neural Networks Using Queries and Counterexamples},
  booktitle={Proceedings of the 35th International Conference on Machine Learning},
  series={Proceedings of Machine Learning Research}, volume={80}, pages={5247--5256}, year={2018},
  publisher={PMLR}, url={https://proceedings.mlr.press/v80/weiss18a.html}
}

@article{jaeger2000observable,
  author={Jaeger, Herbert},
  title={Observable operator models for discrete stochastic time series},
  journal={Neural Computation}, volume={12}, number={6}, pages={1371--1398}, year={2000}, doi={10.1162/089976600300015411}
}

@inproceedings{littman2001predictive,
  author={Littman, Michael L. and Sutton, Richard S. and Singh, Satinder},
  title={Predictive Representations of State},
  booktitle={Advances in Neural Information Processing Systems}, volume={14}, year={2001},
  url={https://proceedings.neurips.cc/paper/2001/hash/1e4d36177d71bbb3558e43af9577d70e-Abstract.html}
}

@article{boots2011closing,
  author={Boots, Byron and Siddiqi, Sajid and Gordon, Geoffrey J.},
  title={Closing the learning-planning loop with predictive state representations},
  journal={The International Journal of Robotics Research}, volume={30}, number={7}, pages={954--966}, year={2011}, doi={10.1177/0278364911404092}
}

@article{olshausen1997sparse,
  author={Olshausen, Bruno A. and Field, David J.},
  title={Sparse coding with an overcomplete basis set: A strategy employed by {V1}?},
  journal={Vision Research}, volume={37}, number={23}, pages={3311--3325}, year={1997}, doi={10.1016/S0042-6989(97)00169-7}
}

@misc{elhage2022superposition,
  author={Elhage, Nelson and Hume, Tristan and Olsson, Catherine and Schiefer, Nicholas and Henighan, Tom and Kravec, Shauna and Hatfield-Dodds, Zac and Lasenby, Robert and Drain, Dawn and Chen, Carol and others},
  title={Toy Models of Superposition}, year={2022}, howpublished={Transformer Circuits Thread},
  url={https://transformer-circuits.pub/2022/toy_model/index.html}, note={Accessed 6 October 2026}
}

@inproceedings{huben2024sparse,
  author={Huben, Robert and Cunningham, Hoagy and Smith, Logan Riggs and Ewart, Aidan and Sharkey, Lee},
  title={Sparse Autoencoders Find Highly Interpretable Features in Language Models},
  booktitle={The Twelfth International Conference on Learning Representations}, year={2024},
  url={https://openreview.net/forum?id=F76bwRSLeK}
}

@inproceedings{shai2024belief,
  author={Shai, Adam S. and Marzen, Sarah E. and Teixeira, Lucas and Gietelink Oldenziel, Alexander and Riechers, Paul M.},
  title={Transformers Represent Belief State Geometry in their Residual Stream},
  booktitle={Advances in Neural Information Processing Systems},
  volume={37},
  pages={75012--75034},
  year={2024},
  doi={10.52202/079017-2387}
}

@inproceedings{tamkin2024codebook,
  author={Tamkin, Alex and Taufeeque, Mohammad and Goodman, Noah D.},
  title={Codebook Features: Sparse and Discrete Interpretability for Neural Networks},
  booktitle={Proceedings of the 41st International Conference on Machine Learning},
  series={Proceedings of Machine Learning Research},
  volume={235},
  pages={47535--47563},
  year={2024},
  publisher={PMLR},
  url={https://proceedings.mlr.press/v235/tamkin24a.html}
}

@inproceedings{torre2025hmm,
  author={Torre, Elia and Viscione, Michele and Pompe, Lucas and Grewe, Benjamin F. and Mante, Valerio},
  title={Mechanistic Interpretability of {RNNs} emulating Hidden Markov Models},
  booktitle={Advances in Neural Information Processing Systems},
  volume={38},
  pages={168883--168926},
  year={2025},
  doi={10.52202/085713-5087}
}

\appendix
\section{HMM emission matrices}\label{app:hmm}

The observability regimes use the same uniform initial state distribution, with probability $0.25$ assigned to each hidden state. Rows correspond to hidden states. Columns correspond to the six visible symbols.

\subsection{Easy}
\begin{equation}
E_{\mathrm{easy}}=\begin{bmatrix}
0.78&0.10&0.03&0.03&0.03&0.03\\
0.03&0.03&0.78&0.06&0.05&0.05\\
0.03&0.03&0.03&0.72&0.10&0.09\\
0.10&0.08&0.08&0.08&0.34&0.32
\end{bmatrix}.
\end{equation}

\subsection{Medium}
\begin{equation}
E_{\mathrm{medium}}=\begin{bmatrix}
0.45&0.35&0.05&0.05&0.05&0.05\\
0.05&0.05&0.65&0.10&0.10&0.05\\
0.05&0.05&0.05&0.35&0.25&0.25\\
0.20&0.15&0.15&0.15&0.20&0.15
\end{bmatrix}.
\end{equation}

\subsection{Hard}
\begin{equation}
E_{\mathrm{hard}}=\begin{bmatrix}
0.24&0.20&0.15&0.14&0.14&0.13\\
0.14&0.14&0.25&0.18&0.16&0.13\\
0.13&0.14&0.14&0.22&0.19&0.18\\
0.18&0.16&0.16&0.16&0.18&0.16
\end{bmatrix}.
\end{equation}

\section{Additional canonical metrics}

Table~\ref{tab:additional} reports held-out state-recovery and donor-swap metrics that supplement the primary results.

\begin{table}[H]
\centering
\caption{Additional held-out canonical metrics}
\label{tab:additional}
\small
\begin{tabular}{lrrrrr}
\toprule
Regime & Bayes ceiling & Cluster acc. & State-probe acc. & Same-swap KL & Different-swap KL \\
\midrule
Easy   & 0.557 & 0.491 & 0.551 & 0.00873 & 0.23562 \\
Medium & 0.517 & 0.432 & 0.512 & 0.00731 & 0.07950 \\
Hard   & 0.380 & 0.291 & 0.370 & 0.00099 & 0.00302 \\
\bottomrule
\end{tabular}
\end{table}

\end{document}